\documentclass[11pt]{article}

\usepackage[preprint]{acl}

\usepackage{times}
\usepackage{latexsym}

\usepackage[T1]{fontenc}

\usepackage[utf8]{inputenc}

\usepackage{microtype}

\usepackage{inconsolata}

\usepackage{graphicx}

\usepackage{graphicx}

\usepackage{amsmath}
\usepackage{amssymb}
\usepackage{mathtools}
\usepackage{amsthm}

\usepackage[most]{tcolorbox}  
\tcbuselibrary{skins}

\usepackage{booktabs}
\usepackage{caption}
\usepackage{makecell}

\usepackage{multirow}
\usepackage[dvipsnames]{xcolor}
\usepackage[table]{xcolor}
\usepackage[normalem]{ulem}
\usepackage{pifont}
\useunder{\uline}{\ul}{}

\usepackage{wrapfig}
\usepackage{lipsum}
\usepackage{array}
\usepackage{tabulary}
\usepackage{tabularx}
\usepackage{listings}
\usepackage{enumitem}
\usepackage{algorithm}
\usepackage{algorithmic}

\definecolor{mygreen}{HTML}{6AA66E}
\definecolor{myblue}{HTML}{5471AB}
\definecolor{myred}{HTML}{B65655}
\definecolor{myorange}{HTML}{D1885C}
\definecolor{mypurple}{HTML}{7F73AF}
\definecolor{mybrown}{HTML}{8F7963}

\newcommand{\up}[1]{$^{\textcolor{RoyalBlue}{+#1}}$} 
\newcommand{\down}[1]{$^{\textcolor{gray}{-#1}}$}

\lstdefinestyle{mystyle}{
    basicstyle=\ttfamily\footnotesize,
    breaklines=true, 
    columns=flexible, 
    captionpos=b,
}
\usepackage{subcaption}

\newcommand{\ie}{\emph{i.e., }}
\newcommand{\eg}{\emph{e.g., }}

\newcommand{\cf}{\emph{cf. }}

\newtcolorbox{paperbox}[1][Summary]{
    enhanced,
    colback=blue!3,
    colframe=NavyBlue,
    attach boxed title to top left={xshift=2mm, yshift=-3mm},
    boxed title style={
        colback=NavyBlue,
        rounded corners, 
        arc=3pt,
        top=0pt, bottom=0pt, left=2pt, right=2pt
    },
    fonttitle=\bfseries\small, 
    title={#1}, 
    top=10pt,
    bottom=6pt,
    left=4pt,
    right=4pt,
    boxsep=1pt,
    arc=3pt,
    boxrule=0.8pt,
    before skip=5pt,
    after skip=5pt,
    drop fuzzy shadow
}

\title{ARMOR: Stabilizing On-Policy LLM RL with Off-Policy Anchor Samples}

\author{
 \textbf{Kexin Huang\textsuperscript{1}},
 \textbf{Junkang Wu\textsuperscript{1}},
 \textbf{Jinda Lu\textsuperscript{1}},
 \textbf{Yang Shuo\textsuperscript{2}},
 \textbf{Chiyu Ma},\\
 \textbf{Jiancan Wu\textsuperscript{1}},
 \textbf{Xiang Wang\textsuperscript{1}},
 \textbf{Xiangnan He\textsuperscript{1}},
 \textbf{Guoyin Wang},
 \textbf{Jingren Zhou}
\\
 \textsuperscript{1}University of Science and Technology of China
 \hspace{10pt}
 \textsuperscript{2}Peking University
\\
\texttt{huangkx@mail.ustc.edu.cn},
\texttt{\{xiangwang1223, xiangnanhe\}@gmail.com}
}

\begin{document}
\maketitle
\begin{abstract}
Reinforcement learning (RL) has significantly enhanced the reasoning capabilities of large language models (LLMs), yet the training process remains notoriously fragile. 
In this work, we investigate a critical source of this instability: \textbf{over-optimization}, where models exploit training heuristics at the expense of generalizable reasoning. 
While reverse KL regularization is the standard defense against such degradation, our analysis reveals that it is often insufficient in this regime, as it fails to ensure comprehensive coverage of the reference distribution. 
To address this, we propose \textbf{ARMOR ({\ul A}nchor {\ul R}ollout and {\ul M}ixed {\ul O}ptimization for {\ul R}L)}, a framework that shifts the paradigm from passive penalty to active sample stabilization. 
ARMOR comprises two key components: 
(1) \textbf{Anchor Rollout}, which leverages off-policy data from the reference policy to preserve established solution patterns; 
and (2) \textbf{Mixed Optimization}, which reformulates the policy objective to enable controlled exploration without relying on auxiliary losses. 
Extensive experiments on reasoning benchmarks validate that ARMOR effectively mitigates validation collapse, enabling sustained performance improvements over extended training horizons.
\end{abstract}

\section{Introduction}

Reinforcement learning has become a central algorithmic driver of recent advances in large language models, substantially enhancing their ability to solve complex tasks and enabling a new class of reasoning-focused models such as OpenAI o1 \citep{Openai-O1}, DeepSeek R1 \citep{Deepseek-R1}, and Qwen3 \citep{Qwen3}. 
Yet, scaling RL for reasoning over long horizons remains fragile: \textbf{stable optimization can still yield unstable generalization}.

\begin{figure}[t]
    \centering
    \begin{subfigure}[b]{0.38\linewidth}
        \centering
        \includegraphics[width=\linewidth]{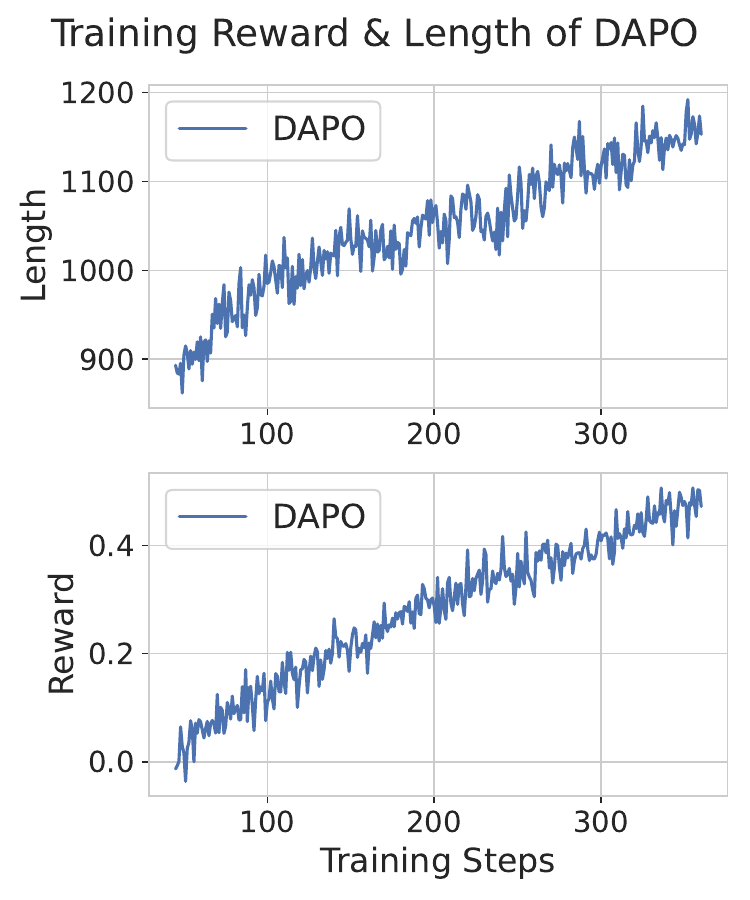}
        \caption{Training behaviors}
        \label{fig:teaser_reward}
    \end{subfigure}
    \hfill
    \begin{subfigure}[b]{0.6\linewidth}
        \centering
        \includegraphics[width=\linewidth]{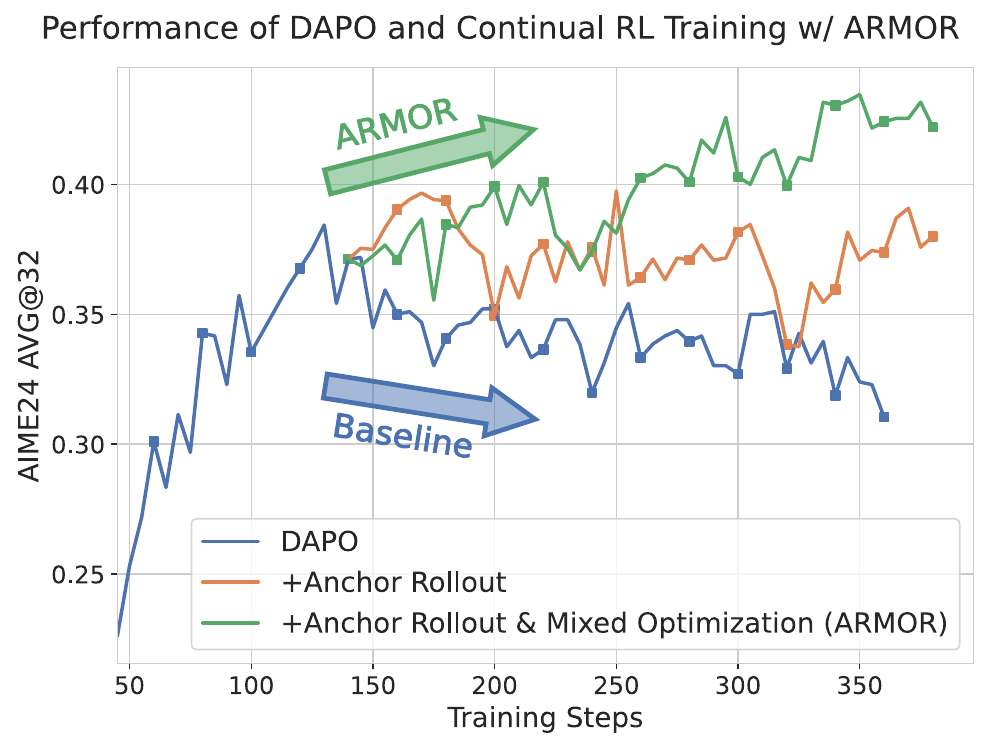}
        \caption{Validation score (AIME24)}
        \label{fig:teaser_perf}
    \end{subfigure}
    \caption{Illustration of the over-optimization issue and the efficacy of ARMOR (on Qwen2.5-Math-7B).
    (a) Training reward and response length steadily rise, indicating stable optimization. 
    (b) However, validation performance reveals a critical generalization failure: the \textcolor{myblue}{baseline} performance degrades after an initial ascent. 
    By integrating \textcolor{myorange}{anchor rollout} to stabilize the sample distribution and mixed optimization to enhance exploration, \textcolor{mygreen}{ARMOR} prevents degradation and sustains continuous performance gains.}
    \label{fig:teaser}
\end{figure}

While the community has extensively addressed system-level instabilities, such as the training-inference mismatch \citep{GSPO, IcePop, XiaomiR3, MiniRL}---an architectural inconsistency that disrupts optimization \citep{TIS, SpeedStability}---we focus on a more fundamental algorithmic failure: \textit{\textbf{over-optimization}} \citep{gao2023scaling}.
As illustrated in Fig.~\ref{fig:teaser}, we observe a distinctive \textit{reward--validation gap}: training reward (answer correctness) improves steadily, yet validation performance decouples and degrades.
In this regime, stable reward curves can even be a warning sign rather than a reassurance.
This phenomenon, echoed in recent works \citep{clipfairly, zhang2025gepo}, indicates a breakdown in generalization rather than an optimization collapse.


Crucially, this failure differs from the classical form of \textit{reward hacking} that arises from a discrepancy between a proxy reward and the true evaluation metric \citep{weng2024rewardhack}.
In reasoning tasks with \textit{verifiable rewards} \citep{tulu-RLVR}, the reward signal remains consistent across training and validation.
Nevertheless, the model still tends to over-optimize solution patterns that yield high rewards on seen data but fail to generalize, necessitating algorithmic intervention beyond simple reward correction \citep{ding2025fapo}.

\begin{figure*}[t]
    \centering
    \begin{subfigure}[b]{0.377\linewidth}
        \centering
        \includegraphics[width=\linewidth]{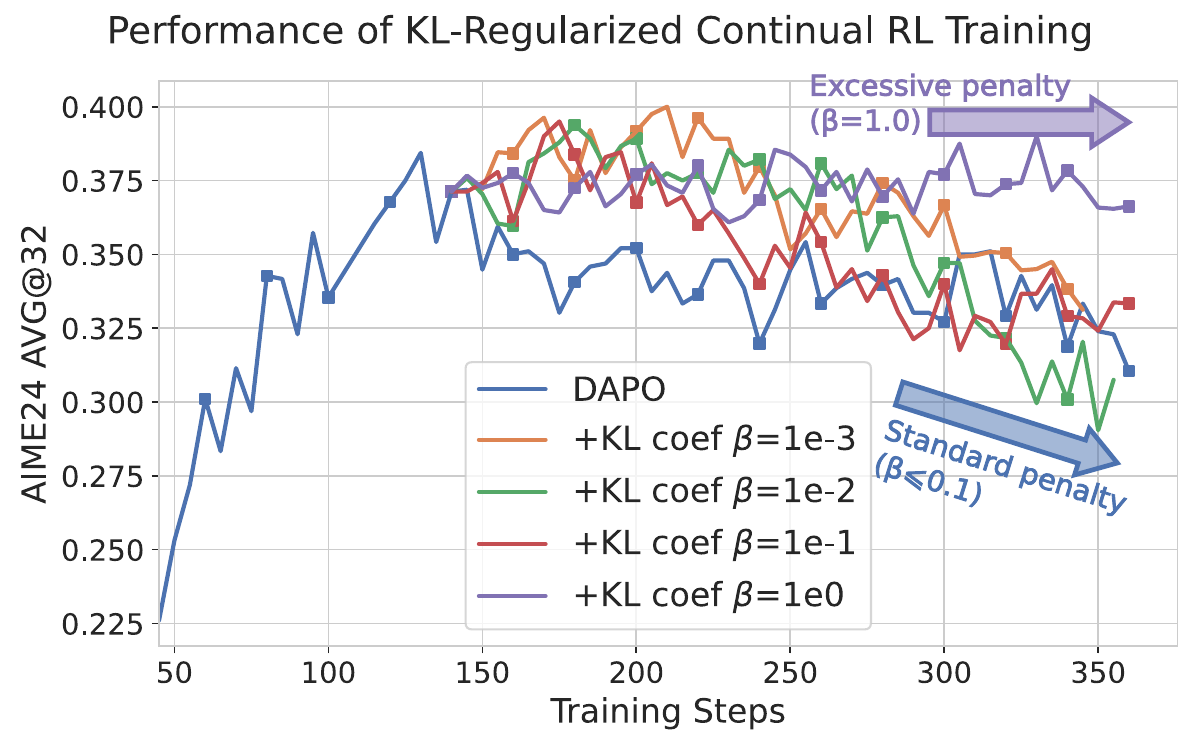}
        \caption{KL regularized validation performance}
        \label{fig:kl_perf}
    \end{subfigure}
    \hfill
    \begin{subfigure}[b]{0.377\linewidth}
        \centering
        \includegraphics[width=\linewidth]{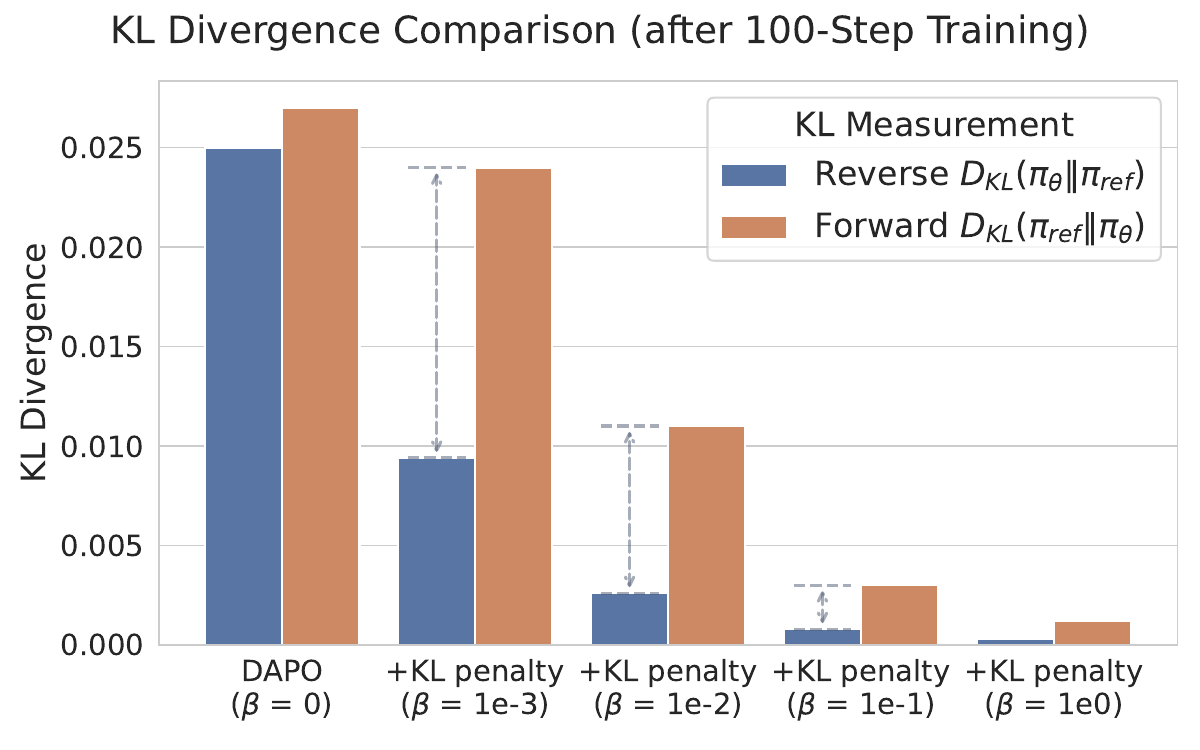}
        \caption{Reverse and forward KL measurements}
        \label{fig:kl_measure}
    \end{subfigure}
    \hfill
    \begin{subfigure}[b]{0.23\linewidth}
        \centering
        \includegraphics[width=\linewidth]{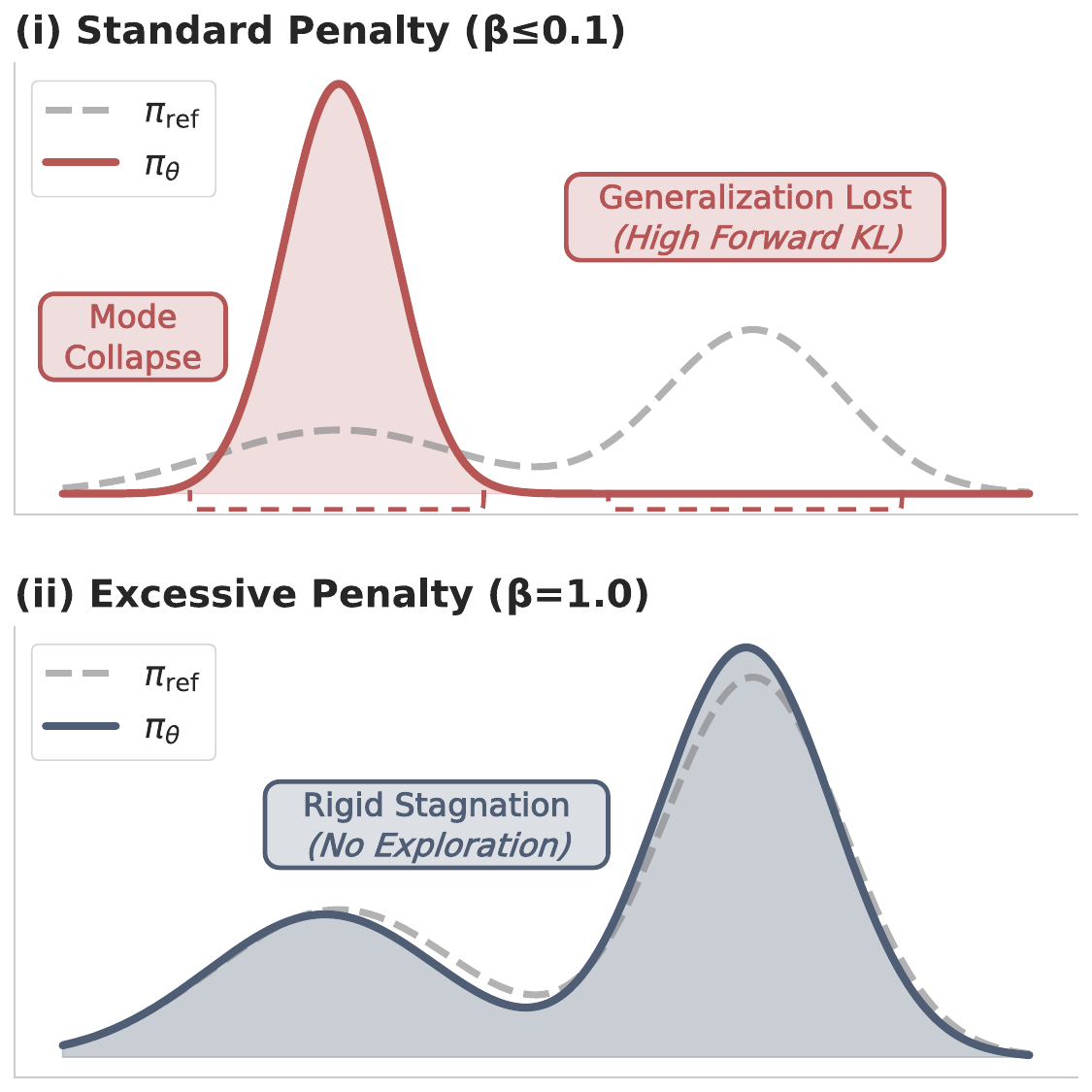}
        \caption{An illustrative example}
        \label{fig:kl_explain}
    \end{subfigure}
    \caption{(a) \textbf{Performance trends under KL regularization. }
    Standard penalties ($\beta \le 0.1$: \textcolor{myorange}{1e-3}, \textcolor{mygreen}{1e-2}, and \textcolor{myred}{1e-1}) merely delay validation collapse of the \textcolor{myblue}{baseline}, while the extreme setting (\textcolor{mypurple}{$\beta$ = 1.0}) ensures stability only by inducing stagnation, highlighting the stability-plasticity dilemma.
    (b) \textbf{Asymmetry in KL divergence} (measured after 100 steps of continual training).
    While increasing $\beta$ drastically reduces the optimized reverse KL $\mathbb D_\text{KL}(\pi_\theta\|\pi_\mathrm{ref})$, it is much less effective at constraining the forward KL $\mathbb D_\text{KL}(\pi_\mathrm{ref}\|\pi_\theta)$.
    (c) \textbf{An illustrative example}. 
    Standard penalties lead to mode collapse (indicated by high Forward KL) due to the mode-seeking nature of the objective, whereas excessive regularization suppresses the exploration required for learning.
    }
\end{figure*}

Unfortunately, we find the standard algorithmic defense, reverse KL regularization \citep{ouyang2022training, gao2023scaling}, is often insufficient in this regime due to two intrinsic limitations (Fig.~\ref{fig:kl_explain}): 
\textbf{(1) Mode-seeking nature (Instability):} Reverse KL only penalizes the policy for generating samples \emph{unlikely} under the reference. Consequently, the policy can still collapse onto a narrow subset of ``shortcut'' patterns without incurring a high penalty, losing the diversity of the reference 
distribution \citep{beyondkl, klcollapse}.
\textbf{(2) Uniform penalty (Stagnation):} By indiscriminately suppressing deviation from the reference, the KL term dampens the exploration required to discover superior reasoning paths that lie beyond the initial distribution.

To address these limitations, we propose \textbf{ARMOR ({\ul A}nchor {\ul R}ollout and {\ul M}ixed {\ul O}ptimization for {\ul R}L)}, a framework that shifts focus from passive loss penalty to active sample stabilization. Our approach comprises two key components, each targeting a specific failure mode of standard KL:
\begin{enumerate}[left=0.5em, topsep=2pt, itemsep=2pt]
\item [(1)] \textbf{Anchor Rollout (Addressing Instability)}: Instead of relying on a passive KL loss to retain distribution modes, we \emph{actively} inject off-policy samples from the reference policy during rollout. 
This acts as an explicit ``anchor'' forcing the model to recall and preserve established generalizable solution patterns.
\item [(2)] \textbf{Mixed optimization (Addressing Stagnation)}: We remove the auxiliary penalty term and instead optimize a mixed policy $\alpha\cdot \pi_\theta + (1-\alpha) \cdot \pi_\mathrm{ref}$, to principally align the optimization target and data distribution.
Crucially, this mixture constructs an adaptive trust region \citep{PPO} that permits controlled exploration without the suppressive effect of a uniform penalty.
\end{enumerate}
As visualized in Fig.~\ref{fig:teaser_perf}, ARMOR effectively bridges the generalization gap, maintaining superior validation performance compared to the degraded baseline.
We validate these benefits through extensive experiments across different base models (\eg Qwen2.5-Math-7B, Qwen3-8B-Base) and RL algorithms (\eg DAPO \citep{DAPO}, QAE \citep{QAE}). 
The results on various reasoning benchmarks confirm that our framework not only secures prolonged training stability but also unlocks continuous performance gains.

\section{Preliminaries}

\textbf{Group Relative Policy Optimization} \citep[\textbf{GRPO},][]{GRPO}.
GRPO removes the separate critic model in Proximal Policy Optimization \citep[PPO,][]{PPO} to enhance training efficiency.
Given a QA pair $(x,a)$ from the dataset $\mathcal D$, it generates a response group $G=\{y_i\}_{i=1}^{|G|}$ using the current policy $\pi_{\theta_\text{old}}$, computes corresponding rewards $\{R_i\}_{i=1}^{|G|}$, and estimates the advantage via:
\begin{equation}\label{eqn:advantage}
    \hat A_{i,t} = \frac{R_i - \mathrm{mean}(\{R_i\}_{i=1}^{|G|})}{\mathrm{std}(\{R_i\}_{i=1}^{|G|})}.
\end{equation}

The policy $\pi_\theta$ is then optimized by maximizing $\mathcal J_\text{GRPO}(\theta)$, defined as:
\begin{gather*}
   \mathcal J =\mathbb E_{
    \mathmakebox[4em][r]{
        \substack{ 
        \hspace{2em}(x,a)\sim\mathcal{D}\\
        \{y_i\}_{i=1}^{|G|} \sim\pi_{\theta_{\text{old}}}(\cdot|x)
        }
    }}\bigg[
     \frac{1}{|G|}\sum_{i=1}^{|G|}\frac{1}{|y_i|}\sum_{t=1}^{|y_i|}\min \Bigl( r_{i,t}(\theta)\hat{A}_{i,t}, \notag \\
    \;\;  \text{clip}\bigl(r_{i,t}(\theta), 1-\epsilon, 1+\epsilon\bigr)\hat{A}_{i,t} \Bigr) - \beta\mathbb D_\mathrm{KL}(\pi_\theta\|\pi_\mathrm{ref}) \bigg],
\end{gather*}
where $r_{i,t}(\theta) = \frac{\pi_{\theta}(y_{i,t}|x, y_{i,<t})}{\pi_{\theta_{\text{old}}}(y_{i,t}|x, y_{i,<t})}$ denotes the importance sampling ratio with $1\pm\epsilon$ being the clipping range, and the KL penalty  $\mathbb D_\mathrm{KL}(\pi_\theta\|\pi_\mathrm{ref})$ regularizes the policy towards the reference policy $\pi_\mathrm{ref}$.

\noindent\textbf{Dynamic Sampling Policy Optimization} \citep[\textbf{DAPO},][]{DAPO}. DAPO is a prominent critic-free RL algorithm that further refines GRPO. 
It employs token-level loss aggregation and introduces the clip-higher mechanism with two clip range of  $\epsilon_\text{low}, \epsilon_\text{high}$, while removing the KL penalty:
\begin{gather}\label{eqn:DAPO}
   \mathcal J_\text{DAPO}(\theta) = \mathbb E_{
    \mathmakebox[4em][r]{
        \substack{ 
        \hspace{2em}(x,a)\sim\mathcal{D}\\
        \{y_i\}_{i=1}^{|G|} \sim\pi_{\theta_{\text{old}}}(\cdot|x)
        }
    }}\bigg[
     \frac{1}{\sum_{i=1}^{|G|} |y_i|} \sum_{i=1}^{|G|} \sum_{t=1}^{|y_i|}  \\
    \min \Bigl( r_{i,t}(\theta)\hat{A}_{i,t}, \text{clip}\bigl(r_{i,t}(\theta), 1-\epsilon_\text{low}, 1+\epsilon_\text{high}\bigr)\hat{A}_{i,t} \Bigr) \bigg],\notag
\end{gather}
Moreover, DAPO employs a dynamic sampling strategy to ensure the response group is not all correct or all wrong:
$$
0 < \left|\{y_i \mid \text{is\_equivalent}(a, y_i)\}\right| < {|G|}
$$
Given its established efficacy, we adopt DAPO as the primary baseline for our empirical analysis.

\noindent\textbf{KL Regularization}. 
Reverse KL regularization serves as the standard defense against over-optimization in RL training for LLMs \citep{klcomedy}.
Unlike early RLHF methods that apply KL as a reward penalty \citep{ouyang2022training}, reasoning tasks often adopt loss-level regularization to avoid penalizing response length \citep{GRPO}. 
We investigate two primary estimators for this term:
(1) The variance-reduced unbiased estimator (\texttt{k3}):
$\mathbb D_\mathrm{KL}(\pi_\theta\|\pi_\mathrm{ref}) = \frac{\pi_\text{ref}(\cdot)}{\pi_\theta(\cdot)}-\log\frac{\pi_\text{ref}(\cdot)}{\pi_\theta(\cdot)}-1$, which is used in GRPO;
and (2) the squared estimator (\texttt{k2}):
$\mathbb D_\mathrm{KL}(\pi_\theta\|\pi_\mathrm{ref}) = \frac12[\log{\pi_\text{ref}(\cdot)}-\log{\pi_\theta(\cdot)]^2}$, which is shown to provide unbiased gradients \citep{klpitfalls}. 
To ensure rigor, we empirically evaluate both estimators to determine if standard KL can bridge the generalization gap, with \texttt{k3} adopted for our main results and \texttt{k2} comparisons deferred to the Appendix.


\section{Empirical Analysis of KL Regularization}\label{sec:kl}

\textbf{Experiment Setup}.
To investigate the efficacy of KL regularization, we adopt a \textbf{continual training} setting focused on the phase where over-optimization emerges.
We first train Qwen2.5-Math-7B base model \citep{Qwen25Math} using DAPO to identify the peak checkpoint $\pi_\theta^*$ immediately prior to validation degradation.
We then resume training from $\pi_\theta^*$ under varying KL coefficients, aiming to turn around the degradation trend observed in the unregularized baseline.
In this phase, we set the reference policy to the current best model ($\pi_\text{ref} \leftarrow \pi_\theta^*$) rather than the initial base model.
This setup aligns with practical long-horizon RL training scenarios \citep{Deepseek-R1}, ensuring that the regularization term discourages deviation from the \textit{best-known} policy rather than enforcing regression to the weaker initial state\footnote{We also explored regularizing against the base model, but this led to rapid performance deterioration.}.


\textbf{Evaluation Protocol}.
We evaluate reasoning performance on the challenging AIME24 benchmark and report \texttt{avg@k}, the average accuracy with $k=32$ responses sampled per question.
Beyond performance metric, we analyze distributional shifts by monitoring both the forward KL divergence $\mathbb D_\text{KL}(\pi_\text{ref}\|\pi_\theta)$ and the reverse KL divergence $\mathbb D_\text{KL}(\pi_\theta\|\pi_\text{ref})$.
Tracking both directions provides a holistic view of the policy drift: while reverse KL serves as the training penalty, forward KL offers deeper insights into whether the model maintains adequate coverage of the reference distribution.

\textbf{Empirical Results}.
As illustrated in Fig.~\ref{fig:kl_perf}, the unregularized baseline (DAPO) suffers from immediate degradation in validation performance as training proceeds.
Introducing KL penalties reveals a fundamental trade-off:

\textit{1) Standard Regime ($\beta \le 0.1$):} 
It is worth noting that standard RL practices typically employ minimal penalties\footnote{DeepSeek-R1 utilizes $\beta$=1e-3 \citep{Deepseek-R1}.}.
Even when increasing $\beta$ to $0.1$, the model eventually succumbs to \textbf{instability} after a brief period of improvement.

\textit{2) Excessive Regime ($\beta = 1.0$):} 
We further increase $\beta$ to $1.0$ as a stress test to probe the limits of regularization.
Although this extreme setting can suppress instability, it leads to \textbf{stagnation}.
This indicates that such stability is achieved only by rigidly tethering the policy, which precludes the exploration necessary for performance gains.



These observations hold consistently across both \texttt{k2} and \texttt{k3} estimators (\cf Appendix~\ref{appendix:add_exp}), indicating that reverse KL is insufficient for stable \emph{and} improving RL training.


\textbf{Analysis: Why KL Fails}.
To understand the mechanics of this failure, we examine the divergence metrics in Fig.~\ref{fig:kl_measure}. 
A clear \textbf{asymmetry} emerges:
while increasing $\beta$ drastically compresses the optimized reverse KL $\mathbb{D}_{\text{KL}}(\pi_\theta\|\pi_\text{ref})$, the forward KL $\mathbb{D}_{\text{KL}}(\pi_\text{ref}\|\pi_\theta)$ \emph{decreases at a much slower rate}, maintaining a disproportionately large gap.
This failure stems from two distinct mechanisms:

\textit{1) Mode-Seeking Nature (Standard Regime).} 
Under standard penalties, the optimization is dominated by the \textbf{mode-seeking nature} of reverse KL \citep{beyondkl,klcollapse}. 
Fundamentally, reverse KL only penalizes the policy for generating samples that are \textit{unlikely} under the reference distribution, but \textit{not} for missing valid regions of the reference support.
Consequently, the model can collapse onto a narrow subset of ``shortcut'' modes without incurring a high penalty, provided these modes exist within the reference distribution (as illustrated in Fig.~\ref{fig:kl_explain}).
This selective collapse discards other generalizable reasoning paths, resulting in the observed performance degradation and a much higher forward KL (which penalizes dropped modes).

\textit{2) Uniform Penalty (Excessive Regime).}
Under extreme penalties (\eg $\beta=1.0$), the regularization acts as a \textbf{uniform penalty} on all policy deviations.
While this forces $\mathbb{D}_{\text{KL}}(\pi_\theta\|\pi_\text{ref}) \to 0$ and thus locally aligns forward and reverse KL \citep{klpitfalls}, it indiscriminately suppresses both harmful degradation and beneficial exploration.
This causes the stagnation observed in our experiments, rendering the ineffective training.

\begin{paperbox}
Standard KL regularization faces a dilemma: loose constraints lead to \textbf{instability} due to its \textit{mode-seeking nature}, while tight constraints result in \textbf{stagnation} due to its \textit{uniform penalty}.
This analysis confirms that a passive penalty alone is structurally insufficient. 
\end{paperbox}

\section{The ARMOR Framework}

Building on our analysis, we introduce ARMOR (Anchor Rollout and Mixed Optimization for RL). 
This framework replaces the passive KL penalty with an active stabilization mechanism comprising two components, each explicitly targeting a failure mode identified in Section~\ref{sec:kl}.

\subsection{Anchor Rollout: Active Mode Retention}
To address the mode-seeking nature of reverse KL, where the model tends to collapse by forgetting diverse solutions present in the reference, we propose \textbf{Anchor Rollout}.
Instead of relying on loss penalties to implicitly retain modes, we \emph{actively} inject reference samples into the training batch.
Specifically, for each query $x$, we construct a hybrid response group $G$ during rollout:
\begin{equation}
\begin{gathered}
y_i \sim \pi_{\theta_{\text{old}}} (\cdot|x), \quad 1 \le i \le |G|-1,\\
y_\text{anc} \sim \pi_{\text{ref}}(\cdot | x), \quad \text{s.t. } R(x, y_\text{anc}) = 1.
\end{gathered}
\end{equation}
Here, the auxiliary response $y_\text{anc}$ serves as an ``anchor sample'' drawn from the reference policy $\pi_{\text{ref}}$ and is guaranteed to be correct\footnote{This positive-only anchor is critical, as including negative samples drove the model \textit{away} from the reference policy, which will lead to increased instability \citep{Deepseek-V3.2}.} ($R=1$) through rejection sampling (Algo.~\ref{alg:armor}, Line~\ref{line:reject}).
By including this anchor in every response group, we provide an explicit signal preventing the policy $\pi_\theta$ from drifting away from known correct solutions. 
As shown in Fig.~\ref{fig:teaser_perf}, \textbf{this simple intervention effectively stabilizes training} where standard KL fails.

From an implementation perspective, our continual training setting initializes $\pi_\mathrm{ref}$ from the best checkpoint of $\pi_\theta$, ensuring identical model architectures.
This design streamlines deployment:
As outlined in Algo.~\ref{alg:armor} (Line~\ref{line:rollout}), we can synchronize the reference parameters $\theta_\text{ref}$ to the inference engine alongside standard updates.
This enables hybrid sampling without the computational overhead of switching between distinct generation engines.

{
\renewcommand{\algorithmicthen}{}
\begin{algorithm}[t]
\begin{algorithmic}[1]
\STATE \textbf{Input:} Initial parameters $\theta^0, \theta_\text{ref} \leftarrow \theta^0$, Dataset $\mathcal{D}$, hyperparameters $B, \mu, \alpha, \tau$.

\FOR{step $t = 1, \dots, T$}
    \STATE \textcolor{gray}{// Phase 1: Anchor Rollout}
    \STATE Initialize batch $\mathcal{B} \leftarrow \emptyset$
    \WHILE{$|\mathcal{B}| < B$}
        \STATE Sample query batch $\mathcal X \sim \mathcal{D}$

        \STATE \textcolor{gray}{// Shared engine by switching parameter}
        \STATE Load $\theta^{t-1}$, sample on-policy groups $\mathcal{Y}_{\text{on}}$
        \STATE Load $\theta_\text{ref}$, sample off-policy groups $\mathcal{Y}_{\text{off}}$\label{line:rollout}
        
        \STATE Compute rewards $\mathcal R_{\text{on}}$ and $\mathcal R_{\text{off}}$

        \STATE \textcolor{gray}{// Rejection sampling}\label{line:reject}
        \FOR{$x \in \mathcal X$}
        \IF{$\text{Var}(\mathcal R_{\text{on}|x}) > 0, \max(\mathcal R_{\text{off}|x})=1$}
            \STATE Let $y_{\text{anc}} = \arg\max_{y\in \mathcal{Y}_{\text{off}|x}} R(x, y)$
            \STATE $G \leftarrow \mathcal{Y}_{\text{on}|x} \cup \{y_{\text{anc}}\}$
            \STATE $\mathcal{B} \leftarrow \mathcal{B} \cup \{(x,G)\}$ 
        \ENDIF
        \ENDFOR
    \ENDWHILE
    
    \STATE \textcolor{gray}{// Phase 2: Mixed Optimization}
    \FOR{iteration = $1,\dots,\mu$}
    \STATE Update $\theta^t$ by maximizing $\mathcal{J}(\theta)$, using the mixed IS ratio $r_{i,t}^\text{mix}$ (Eq.~\ref{eqn:mixed_is})
    \ENDFOR
    
    \STATE \textcolor{gray}{// (Optional) Reference Reset}\label{line:ref}
    \IF{$t \mod \tau = 0$}
        \STATE $\theta_\text{ref} \leftarrow \theta^t$
    \ENDIF
\ENDFOR
\end{algorithmic}
\caption{ARMOR Training Framework}\label{alg:armor}
\end{algorithm}
}

\subsection{Mixed Optimization: Performance Ceiling}
While Anchor Rollout guarantees stability, our objective extends beyond collapse prevention to maximizing the model's asymptotic performance.
To achieve this, we first revisit the theoretical implications of our rollout strategy, observing that by injecting anchor samples, Anchor Rollout implicitly constructs \textbf{mixture policies} regularized by the reference:
\begin{equation}
\begin{gathered}
\pi^{\text{mix}}_\theta = \alpha\pi_\theta + (1-\alpha)\pi_{\text{ref}},\\
\pi^{\text{mix}}_{\theta_\text{old}} = \alpha\pi_{\theta_\text{old}} + (1-\alpha)\pi_{\text{ref}}
\end{gathered}
\end{equation}
where $\alpha \in (0, 1)$ is the mixing coefficient.
Since the data is generated from this mixture, the \textbf{principled optimization objective} should align with this structure.
Therefore, rather than optimizing $\pi_\theta$ in isolation, we reformulate the objective to optimize the mixture policy itself.

We implement this by substituting the standard probability terms in the policy gradient (e.g., DAPO) with their mixed variants, yielding the mixed Importance Sampling (IS) ratio:
\begin{equation}\label{eqn:mixed_is}
r_{i,t}^\text{mix}(\theta) = \frac{\alpha\pi_\theta(\cdot) + (1-\alpha)\pi_{\text{ref}}(\cdot)}{\alpha\pi_{\theta_{\text{old}}}(\cdot) + (1-\alpha)\pi_{\text{ref}}(\cdot)}.
\end{equation}
This formulation explicitly accounts for the reference policy's contribution to the data distribution.
Crucially, since $\pi_\text{ref}$ is fixed within each step, any performance gain achieved on the mixture target $\pi^{\text{mix}}_\theta$ strictly translates to improvements in the target policy $\pi_\theta$:
\begin{align*}
&\quad\mathbb E_{y\sim\pi_{\theta}^\text{mix}(\cdot|x), y'\sim\pi_{\theta_\text{old}}^\text{mix}(\cdot|x)}
[R(x,y) - R(x,y')] > 0 \\
&\Rightarrow \mathbb E_{y\sim\pi_{\theta}(\cdot|x), y'\sim\pi_{\theta_\text{old}(\cdot|x)}}
[R(x,y) - R(x,y')] > 0.
\end{align*}

Furthermore, we follow the iterative updating paradigm \citep{Deepseek-R1} by periodically resetting $\pi_\text{ref}$ to the current policy state (Algo.~\ref{alg:armor}, Line~\ref{line:ref}), avoiding saturation due to a fixed anchor.


We designate this holistic approach---integrating the stabilization of Anchor Rollout with Mixed Optimization that calibrates the IS ratio---as \textbf{ARMOR}. 
As demonstrated in Fig.~\ref{fig:teaser_perf}, this combination yields a \textbf{significantly higher performance ceiling}.

\subsection{Explanation: Adaptive Trust Region}\label{sec:theory}
To further elucidate how Mixed Optimization enables superior asymptotic performance, we analyze its impact on policy update dynamics through the lens of the trust-region.

PPO variants (\eg GRPO, DAPO) clip the positive/negative sample's IS ratio into fixed intervals $[0, 1+\epsilon]$ or $[1-\epsilon,\infty)$ to enforce a trust region $\pi_\theta \approx \pi_{\theta_\text{old}}$. 
However, substituting our mixed ratio yields a flexible boundary.
Considering the clipping boundaries $r_{i,t}^\text{mix}(\theta) = 1 \pm \epsilon$, we have:
$$
\frac{\alpha\cdot\pi_\theta + (1-\alpha)\cdot\pi_{\text{ref}}}{\alpha\cdot\pi_{\theta_{\text{old}}} + (1-\alpha)\cdot\pi_{\text{ref}}} = 1 \pm \epsilon.
$$
Solving for the current policy $\pi_\theta$, we derive the \textbf{effective trust region boundary} \citep{TRPO,PPO}:
\begin{equation}\label{eqn:clip_expand}
\pi_\theta = \underbrace{(1 \pm \epsilon)
\cdot\pi_{\theta_{\text{old}}}}_{\text{Standard Boundary}} \pm \underbrace{\epsilon (1-\alpha)/{\alpha}\cdot\pi_{\text{ref}}}_{\text{Expansion Term }(\ge 0)}.
\end{equation}
This derivation reveals a critical mechanism: In contrast to the standard clip boundary $\pi_\theta =(1 \pm \epsilon)\cdot\pi_{\theta_{\text{old}}}$, ARMOR \textbf{dynamically expands the clipping range} proportional to the reference probability $\pi_{\text{ref}}$.
This derivation reveals that ARMOR dynamically modulates the update magnitude based on the reference prior, enabling a dual mechanism for performance breakthroughs:

\textit{1) Reinforcing verified correctness:} For positive updates ($1+\epsilon$), the higher upper bound allows larger steps towards correct actions supported by the reference. 
This is crucial for effectively learning from \emph{off-policy} correct samples, ensuring they are not unfairly clipped due to distribution shift.

\textit{2) Rectifying reference biases:} For negative updates ($1-\epsilon$), the lowered boundary permits stronger penalization of incorrect actions, even if they are highly probable in $\pi_{\text{ref}}$. 
This mechanism is key to \emph{surpassing} the reference capability, as it allows the model to decisively correct the reference's inherent biases rather than blindly imitating them.

\begin{table*}[t]
\centering
\caption{Comparison of RL baselines (DAPO, QAE) against their ARMOR continual-training counterparts across Qwen2.5-Math-7B and Qwen3-8B-Base.
ARMOR consistently improves mathematical reasoning (\texttt{avg@32}) on AIME and AMC benchmarks while maintaining comparable general capabilities (\texttt{avg@8} on GPQA and MMLU-Pro), effectively mitigating the over-optimization observed in baselines.}
\label{tab:main}
\resizebox{\linewidth}{!}{
\begin{tabular}{@{}ccc|llll|ll@{}}
\toprule
\multirow{2}{*}{\textbf{Model}} & \multirow{2}{*}{\textbf{Method}} & \multirow{2}{*}{\textbf{Step}} & \multicolumn{4}{c|}{\textbf{Math Reasoning Tasks}} & \multicolumn{2}{c}{\textbf{General Tasks}} \\ \cmidrule(l){4-9} 
 &  &  & \multicolumn{1}{c}{\textbf{AIME24}} & \multicolumn{1}{c}{\textbf{AIME25}} & \multicolumn{1}{c}{\textbf{AMC}} & \multicolumn{1}{c|}{\textbf{Average}} & \multicolumn{1}{c}{\textbf{GPQA}} & \multicolumn{1}{c}{\textbf{MMLU-Pro}} \\ \midrule
\multirow{4}{*}{\begin{tabular}[c]{@{}c@{}}Qwen2.5-\\ Math-7B\end{tabular}} & DAPO & 140 & 37.13 & 15.21 & 69.39 & 40.58 & 38.26 & 43.93 \\
 & \textbf{+ARMOR} & +200 & 43.04\up{5.91} & 18.13\up{2.92} & 76.13\up{6.74} & 45.77\up{5.19} & 42.49\up{4.23} & 46.03\up{2.1} \\ \cmidrule(l){2-9} 
 & QAE & 230 & 39.79 & 15.96 & 73.53 & 43.09 & 40.78 & 43.30 \\
 & \textbf{+ARMOR} & +240 & 41.58\up{1.79} & 16.15\up{0.19} & 76.62\up{3.09} & 44.78\up{1.69} & 39.02\down{1.76} & 42.15\down{1.15} \\ \midrule
\multirow{2}{*}{\begin{tabular}[c]{@{}c@{}}Qwen3-8B-\\ Base\end{tabular}} & DAPO & 70 & 36.98 & 28.13 & 71.72 & 45.61 & 49.87 & 65.55 \\
 & \textbf{+ARMOR} & +220 & 48.13\up{11.15} & 34.48\up{6.35} & 80.35\up{8.63} & 54.32\up{8.71} & 56.25\up{6.38} & 68.90\up{3.35} \\ \bottomrule
\end{tabular}
}
\end{table*}

\begin{paperbox}
ARMOR resolves the stability-exploration dilemma by decoupling the constraints: 
\textbf{Anchor Rollout} prevents \textbf{instability} by actively retaining correct reference solutions, while \textbf{Mixed Optimization} avoids \textbf{stagnation} by constructing an adaptive trust region. 
Empirically, ARMOR secures continuous reasoning improvements without the fragility of standard KL regularization.
\end{paperbox}

\section{Experiments}

In this section, we provide a comprehensive empirical evaluation of ARMOR.
We first present the main results across different base models and RL algorithms to validate the framework's broad effectiveness in mitigating over-optimization. 
Subsequently, we conduct detailed ablation studies to dissect the individual contributions of the two core components in ARMOR, as well as the impact of various reference policy settings.

\subsection{Main Results}

\textbf{Experimental Setup}.
To verify that ARMOR effectively addresses the over-optimization problem, we adopt the \textbf{continual training} setting described in Section~\ref{sec:kl}.
We employ two base models: Qwen2.5-Math-7B \citep{Qwen25Math} and Qwen3-8B-Base \citep{Qwen3}, trained on the DAPO-Math-17K dataset.
For the underlying RL algorithms, we select DAPO \citep{DAPO} and QAE \citep{QAE}. 
QAE represents a state-of-the-art method with a quantile-based advantage function, which we integrate into our framework by replacing the advantage term in Eq.~\eqref{eqn:advantage}.

The training protocol proceeds in two stages: we first run the baseline algorithm until validation performance degrades due to over-optimization. 
We then apply ARMOR to continue training from the best checkpoint (identified via the AIME24 validation set) prior to performance collapse. 
Detailed experimental setting and hyperparameters are provided in Appendix~\ref{appendix:impl}.

\textbf{Benchmarks and Metrics}. 
We evaluate performance on both mathematical reasoning and general capabilities. 
For reasoning, we report both the average accuracy (\texttt{avg@k}) and \texttt{pass@k} \citep{passk} on challenging mathematical benchmarks, including AIME24/25 and AMC. 
We also evaluate the model's general capabilities on science reasoning \citep[GPQA,][]{GPQA} and language understanding \citep[MMLU-Pro,][]{MMLU-Pro}. 
For math reasoning tasks, we sample 32 responses to derive a robust estimation. 
For general tasks, we use 8 samples due to the significantly larger query volume. 
All reported checkpoints are selected based on their AIME24 performance.

\begin{figure*}[t]
    \centering
    \begin{subfigure}[b]{0.325\linewidth}
        \centering
        \includegraphics[width=\linewidth]{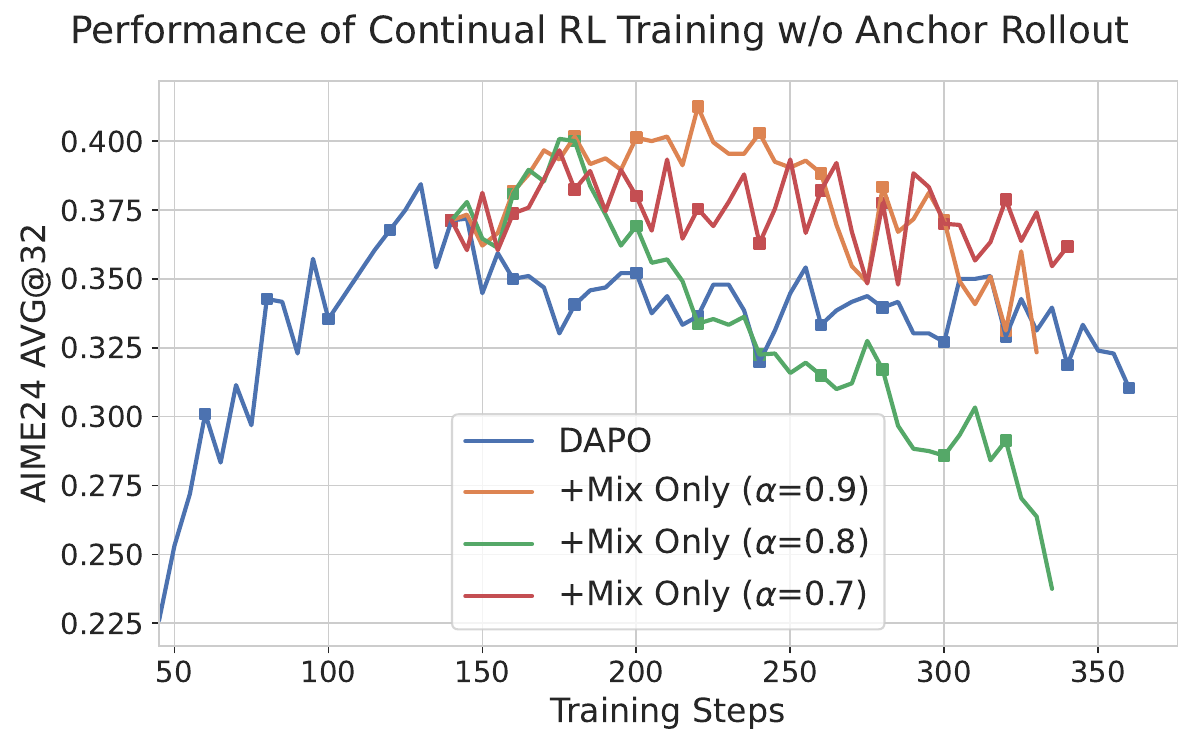}
        \caption{Without Anchor Rollout}
        \label{fig:no_anchor}
    \end{subfigure}
    \hfill
    \begin{subfigure}[b]{0.325\linewidth}
        \centering
        \includegraphics[width=\linewidth]{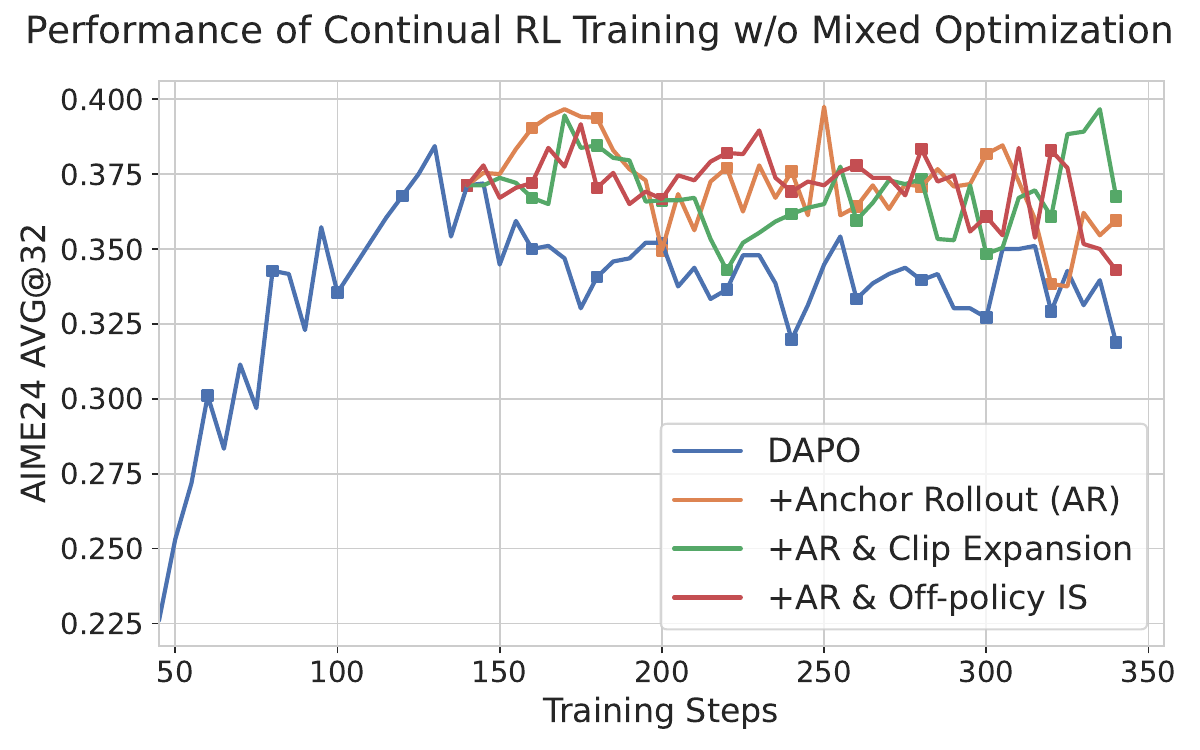}
        \caption{Without Mixed Optimization}
        \label{fig:no_mixed}
    \end{subfigure}
    \hfill
    \begin{subfigure}[b]{0.325\linewidth}
        \centering
        \includegraphics[width=\linewidth]{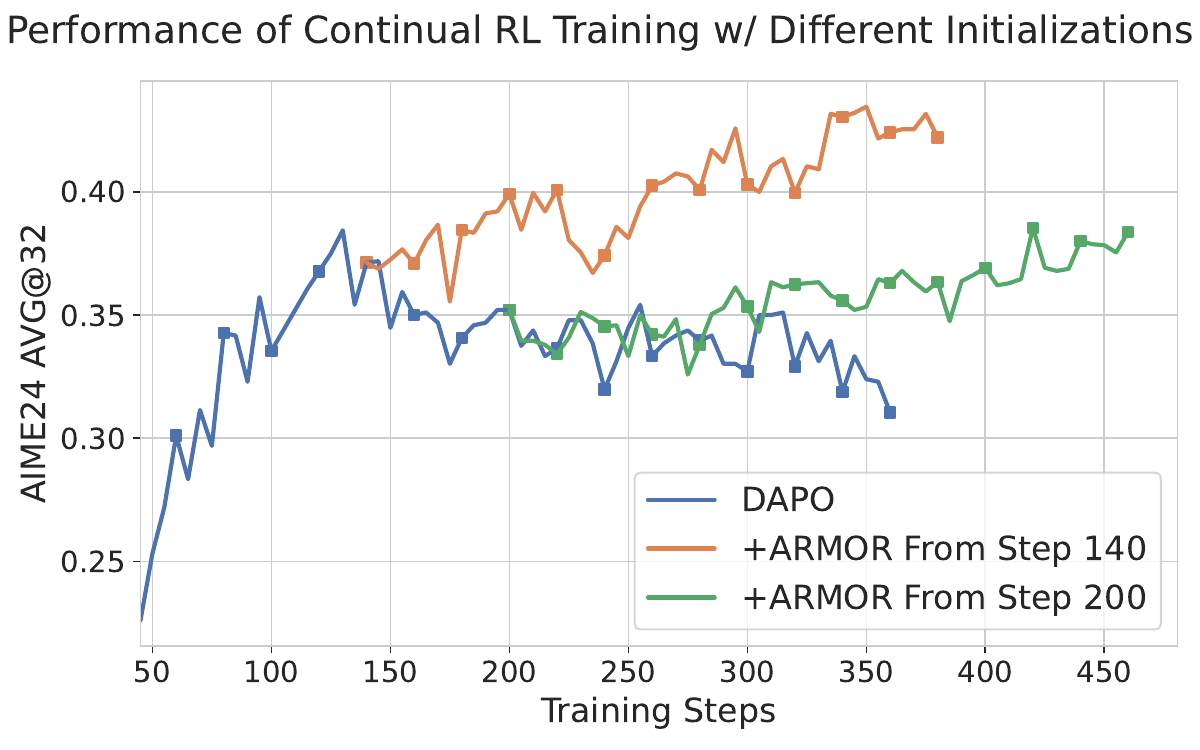}
        \caption{Robustness to Initial Model Quality}
        \label{fig:diff_start}
    \end{subfigure}
    \caption{(a) \textbf{Necessity of Anchor Rollout}. 
    We compare \textcolor{myblue}{DAPO} against variants using \emph{only} Mixed Optimization ($\alpha$ = \textcolor{myorange}{0.9}, \textcolor{mygreen}{0.8}, and \textcolor{myred}{0.7}).
    Without the stability foundation provided by Anchor Rollout, these models suffer from eventual performance collapse.
    (b) \textbf{Necessity of Mixed Optimization}.
    We replace mixed optimization with alternative objectives on top of Anchor Rollout (AR).
    While \textcolor{myorange}{AR alone} effectively stabilizes training, modifying standard objectives with \textcolor{mygreen}{comparable clip range} or \textcolor{myred}{off-policy IS} fails to yield further improvements.
    (c) \textbf{Ablation of different initializations}.
    We apply ARMOR to models initialized from an optimal checkpoint \textcolor{myorange}{(Step 140)} versus a collapsed checkpoint \textcolor{mygreen}{(Step 200)}. 
    Remarkably, ARMOR successfully recovers performance even from a degraded state, although initiating from a higher-quality model enables a higher final performance peak.
    }
\end{figure*}

\textbf{Performance Analysis}.
As summarized in Tab.~\ref{tab:main}, while standard training suffers from over-optimization (so the select checkpoints undergo fewer training steps), ARMOR successfully stabilizes the training process and secures sustained performance gains beyond the initial peak:
\begin{itemize}[left=0em,topsep=0pt,itemsep=0pt]
    \item \textbf{Significant Gains with DAPO:} On both Qwen2.5 and Qwen3 variants, applying ARMOR to DAPO yields substantial improvements, boosting average math scores by \textbf{+5--8 points}. Notably, this reasoning gain does not come at the expense of general capabilities, which also see clear enhancement by \textbf{+2--6 points}.
    \item \textbf{Robustness with QAE:} When applied to QAE, ARMOR further improves mathematical reasoning by \textbf{+1.7 points}, validating its robustness across different algorithms.
    While we observe a slight regression in general capabilities, we attribute this to two factors: the extended training horizon naturally increases the risk of forgetting for general knowledge, and more importantly, QAE's advantage masking mechanism\footnote{QAE uses reward quantiles as an advantage baseline, naturally producing zero-advantage samples (typically $\sim$80\% of the batch).} can inadvertently discard our injected anchor samples $y_\text{anc}$, thereby weakening the regularization effect (We also observe a larger $\mathbb D_\text{KL}(\pi_\text{ref}\|\pi_\theta)$ than DAPO's in our experiments).
\end{itemize}
Overall, ARMOR consistently extends the effective training window and pushes the reasoning performance ceiling across all tested settings.
As a two-stage recipe, ARMOR also outperforms teacher-guided OPD~\citep{agarwal2024policy} and MOPD~\citep{xiao2026mimo} baselines initialized from the same best zero-RL checkpoint (Appendix~\ref{appendix:two_stage}).

\begin{table}[t]
\centering
\caption{
\texttt{Pass@k} evaluation of ARMOR compared to baselines. 
The checkpoints used for evaluation are consistent with Table~\ref{tab:main}. 
ARMOR maintains or improves \texttt{pass@16} despite extended training horizons, indicating robust intrinsic reasoning improvements.
}
\label{tab:passk}
\resizebox{\linewidth}{!}{
\begin{tabular}{@{}ccccc@{}}
\toprule
\textbf{} & \textbf{AIME24} & \textbf{AIME25} & \textbf{AMC} & \textbf{Average} \\ \midrule
7B-DAPO & 55.75 & 34.96 & 90.82 & 60.51 \\
\textbf{+ARMOR} & 56.52 & 39.44 & 90.89 & \textbf{62.28} \\ \midrule
7B-QAE & 51.21 & 30.71 & 88.52 & 56.81 \\
\textbf{+ARMOR} & 50.33 & 34.08 & 86.12 & \textbf{56.84} \\ \midrule
8B-DAPO & 70.65 & 50.88 & 90.24 & 70.59 \\
\textbf{+ARMOR} & 74.60 & 51.42 & 90.66 & \textbf{72.23} \\ \bottomrule
\end{tabular}
}
\end{table}

\textbf{\texttt{Pass@k} Evaluation}.
Beyond average performance, we investigate intrinsic reasoning capability via the \texttt{pass@k} metric \cite{yue2025does}.
Tab.~\ref{tab:passk} reports \texttt{pass@16} results on three math benchmarks \citep{passk}. 
ARMOR achieves comparable or improved scores across all models and baselines, confirming that our method effectively raises the model's absolute problem-solving ceiling rather than simply trading off \texttt{pass@k} for accuracy.

\vspace{-3pt}
\subsection{Ablation Study}
To understand the source of ARMOR's gains, we dissect the framework to analyze the necessity of Anchor Rollout, Mixed Optimization, and the impact of reference policy.

\begin{figure*}[t]
    \centering
    \includegraphics[width=\linewidth]{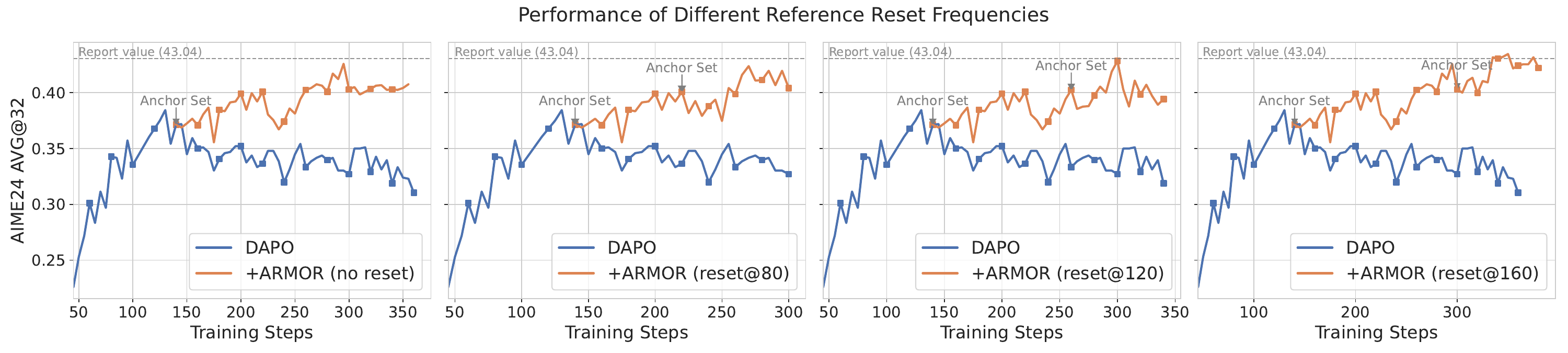}
    \caption{\textbf{Impact of Reference Reset Frequency.} 
    We compare ARMOR training with a static anchor (no reset) versus resetting the reference policy every 80, 120, or 160 steps (noted by ``Anchor Set'' in the figure).
    Periodic resets allows the anchor to evolve with the policy, preventing stagnation and fostering higher reasoning accuracy.}
    \label{fig:diff_reset}
\end{figure*}

\vspace{-3pt}
\subsubsection{Necessity of Anchor Rollout}
We first test a variant using only Mixed Optimization (Eq.~\ref{eqn:mixed_is}) without injecting anchor samples, so the only difference from DAPO here is the replaced IS ratio: $r_{i,t}^\text{mix}$.
As shown in Fig.~\ref{fig:no_anchor}, on the one hand, this variant can achieve initial growth across different mixing coefficients $\alpha$, with the peak performance increased from 37 to 40+, showcasing Mixed Optimization's effect on improving the performance ceiling.
On the other hand, without the explicit retention provided by anchor samples, the model eventually degenerates.

This result demonstrates that the exploration capability provided by Mixed Optimization alone is insufficient for stable RL scaling; \textbf{Anchor Rollout is essential for providing the stability floor} required to prevent collapse.

\subsubsection{Necessity of Mixed Optimization}\label{sec:mix_ablation}
Next, we remove Mixed Optimization and evaluate Anchor Rollout paired with three alternative optimization objectives:
\textit{(a) Standard DAPO}: default objective with no modification to the IS ratio;
\textit{(b) Static Clip Expansion}: DAPO with a manually expanded clipping boundary $1\pm\epsilon/\alpha$, designed to match the theoretical effective clip boundary of Mixed Optimization;
\textit{(c) Standard Off-Policy IS}: Setting IS ratio based on the sampling distribution:
$$
r_{i,t}^\text{off} = \left\{
\begin{aligned}
&{\pi_\theta}(y_{i,t}|x)/{\pi_{\theta_{old}}(y_{i,t}|x)} && 1\le i \le |G|-1,\\
&{\pi_\theta(y_{i,t}|x)}/{\pi_\mathrm{ref}(y_{i,t}|x)} && i=|G|.
\end{aligned}
\right.
$$
Note that we include (b) to ensure a comprehensive ablation:
As derived in Sec.~\ref{sec:theory}, the clip boundary of mixed optimization is \textit{dynamically} broadened to $(1\pm\epsilon)\pi_{\theta_\text{old}}\pm\epsilon(1-\alpha)/\alpha\cdot\pi_\text{ref}$.
If we were to replace $\pi_\text{ref}$ with $\pi_{\theta_\text{old}}$, this boundary would simplify to $(1\pm\epsilon/\alpha)\pi_{\theta_\text{old}}$.
To verify that the gain of Mixed Optimization comes from the specific inclusion of $\pi_\text{ref}$ rather than simply a wider clip range, we test this \textit{static} expanded clip boundary of $1\pm\epsilon/\alpha$ in baseline (b).
We also ensure that the hyper-parameter $\alpha$ here is identical to the one used in our ARMOR experiments for a fair comparison.

As illustrated in Fig.~\ref{fig:no_mixed}, while all three alternatives successfully stabilize training (thanks to Anchor Rollout), they hit a significantly lower performance ceiling compared to the full ARMOR framework.
Baseline (b)'s failure specifically highlights that simple expansion is insufficient; the gain comes from the \textit{adaptive trust region} that selectively reinforces verified correctness and rectifies reference biases.
This demonstrates that ARMOR’s advantage is two-fold: \textbf{Anchor Rollout ensures stability, while Mixed Optimization unlocks the exploration capability required to reach a higher ceiling.}

\subsubsection{Impact of Reference Policy}
Finally, we analyze the impact of the reference policy, focusing on reset frequency and anchor quality.

\textit{1) Impact of Reset Frequency.}
We compared training without resets against resetting the anchor after 80, 120, and 160 steps (Line~\ref{line:ref} in Algo.~\ref{alg:armor}). 
As shown in Fig.~\ref{fig:diff_reset}, while all ARMOR variants outperform the baseline, periodic resets consistently achieve a higher performance ceiling than no reset. 
This confirms that the anchor must evolve alongside the model to avoid becoming a bottleneck.

\textit{2) Robustness to Initial Anchor Quality.}
We tested initializing ARMOR from two distinct points: a \textit{Peak Checkpoint} (step 140, optimal performance) and a \textit{Degenerated Checkpoint} (step 200, after performance drop). Both settings underwent a consistent reset frequency of 160 for a fair comparison.
Remarkably, ARMOR yielded clear improvements in both cases (Fig.~\ref{fig:diff_start}), demonstrating its ability to recover a failing model. However, starting from the higher-quality anchor allowed the model to reach a markedly higher final ceiling.
These results emphasize that while ARMOR is robust, a well-timed application strategy is critical for maximizing ultimate reasoning capabilities.

\section{Conclusion}
In this work, we tackle the critical challenge of \textit{over-optimization} in scaling RL for reasoning.
We identify that standard reverse KL regularization is insufficient due to an intrinsic \textbf{stability-exploration dilemma}: its mode-seeking nature leads to mode collapse, while uniform constraints cause stagnation.
To resolve this, we introduce \textbf{ARMOR}, a framework that synergizes \textbf{Anchor Rollout} to actively stabilize the sample distribution and \textbf{Mixed Optimization} to construct an adaptive trust region for exploration.
Extensive experiments confirm that ARMOR effectively prevents degradation and unlocks sustained performance gains across diverse training settings.

\newpage
\section*{Limitations}
While ARMOR offers a robust solution for RL scaling, we acknowledge several limitations that point toward future directions:
\textit{(1) Computational Overhead:} Incorporating anchor samples through rejection sampling incurs additional inference cost.
While we regard this as a justified investment for stable scaling, future work could mitigate this cost with an offline replay buffer to recycle high-quality anchors.
\textit{(2) Additional Hyperparameter:} Currently, the reference reset strategy is determined manually. 
Future work could explore adaptive scheduling to automate this process.
\textit{(3) Theoretical Analysis:} While our trust-region analysis explains how Mixed Optimization safely incorporates more signals, establishing a theoretical guarantee linking this adaptation to superior convergence or optimality remains an open challenge.

\bibliography{bibilography}

\newpage
\appendix

\section{Related Work}
\paragraph{Reinforcement learning for LLM.}
While early foundational RL works in LLM post-training centered on \textit{Reinforcement Learning from Human Feedback} (RLHF) to align models with human preferences \citep{rlhf-stiennon2020learning, ouyang2022training}, the RL paradigm has recently shifted towards \textit{Reinforcement Learning with Verifiable Rewards} (RLVR) for verifiable tasks such as mathematics and coding \citep{tulu-RLVR}.
Under this regime, OpenAI o1 \citep{Openai-O1} is a seminal reasoning-oriented LLM built by RL, while DeepSeek-R1 \citep{Deepseek-R1} introduces a detailed recipe using GRPO \citep{GRPO}.
These milestones have catalyzed a surge of research dedicated to reproducing reasoning behaviors \citep{ORZ, DAPO}, investigating underlying mechanisms such as entropy dynamics \citep{yue2025does, entropy-mechanism, entropy-80, huang2026direction, meng2026sparse}, and refining learning algorithms \citep{Dr.GRPO, QAE, GMPO, ma2026fipo}.
In this work, we adopt DAPO \citep{DAPO} as our primary backbone.


\paragraph{Stability Issues in LLM RL.}
Despite these advancements, scaling RL training faces significant stability challenges:
\textit{(1) System-level Instability.} 
RL frameworks typically decouple rollout (\eg vllm \citep{vllm}) and training engines (\eg Megatron \citep{megatron-lm}) to maximize throughput, introducing a \textit{training-inference mismatch} \citep{TIS, SpeedStability}.
This architectural inconsistency, which is exacerbated in Mix-of-Expert (MoE) models by diverging router behaviors \citep{GSPO}, can destabilize optimization.
Although extensive existing works address this problem \citep{IcePop, XiaomiR3, MiniRL}, stable optimization can still yield unstable generalization, and we focus on this fundamental challenge:
\textit{(2) Algorithmic Over-optimization.}
This issue arises when stable improvement in training rewards paradoxically leads to degraded generalization \citep{gao2023scaling, repo}.
Unlike classic \textit{Reward Hacking} driven by proxy reward flaws \citep{weng2024rewardhack}, this phenomenon persists in RLVR even when reward signals are consistent with evaluation.
While some studies attribute this to ``false positive'' reasoning \citep{why-hallucinate, ding2025fapo}, we argue that reward shaping might not be the sole cure.
Given that imperfect outcome rewards remain highly effective \citep{ding2025fapo, yang2026clipping} and evaluation metrics mirror training objectives, the degradation implies a fundamental \textbf{algorithmic tendency toward over-optimization}.
Thus, we advocate for algorithmic interventions to enforce stability, serving as a necessary foundation that complements reward modeling.

\paragraph{Learning with off-policy guidance.}
Using off-policy data is a common strategy in RL, typically serving two purposes: efficiency and distillation.
For example, methods like Experience Replay \citep{repo_replay, remix, exgrpo} and unified SFT-RL frameworks \citep{luffy, upt, mentor} utilize historical or expert trajectories alongside on-policy generations for improved efficiency or better performance.
In addition, recent works \citep{rlep, yang2026one, lu2026experience} also explored a two-stage training paradigm, where an optimized RL model guides a base model to bootstrap capabilities.
While ARMOR also utilizes off-policy samples $y_\text{anc}$ from $\pi_\text{ref}$ and employs a two-stage training paradigm, it differs fundamentally in its target:
We utilize the reference not to bootstrap a new model, but to serve as a dynamic anchor that sustains the model's own scaling and explicitly counteracts the tendency toward over-optimization.\\

\paragraph{KL Regularization and Divergence Choices.}
KL regularization is widely used to constrain policy updates and mitigate reward over-optimization in RLHF/RLVR \citep{ouyang2022training, gao2023scaling, klcomedy}. 
However, reverse KL is mode-seeking and can still allow the policy to collapse onto a narrow subset of high-reward patterns \citep{beyondkl, klcollapse}, motivating recent studies of alternative divergences, including forward-KL-style exploration in RAPO \citep{deng2025unlocking} and f-divergence-based GRPO variants \citep{li2025choice}. 
Different from existing RLVR works that mainly use alternative divergences to preserve diversity or enhance exploration (pass@k), our work targets late-stage \textbf{over-optimization}, where training reward keeps increasing while validation accuracy collapses, and addresses it through Anchor Rollout and Mixed Optimization.
\begin{table*}[t]
\centering
\caption{
Supplementary comparison with teacher-guided two-stage training.
OPD and MOPD use the same DAPO-trained checkpoint as a teacher policy, while ARMOR continues training from this checkpoint to address over-optimization.
All checkpoints are selected by AIME24 performance, and math results report \texttt{avg@32}.
}
\label{tab:two_stage}
\resizebox{\linewidth}{!}{
\begin{tabular}{@{}cc|cccc|cc@{}}
\toprule
\multirow{2}{*}{\textbf{Model}} & \multirow{2}{*}{\textbf{Method}} & \multicolumn{4}{c|}{\textbf{Math Reasoning Tasks}} & \multicolumn{2}{c}{\textbf{General Tasks}} \\ \cmidrule(l){3-8}
 & & \multicolumn{1}{c}{\textbf{AIME24}} & \multicolumn{1}{c}{\textbf{AIME25}} & \multicolumn{1}{c}{\textbf{AMC}} & \multicolumn{1}{c|}{\textbf{Average}} & \multicolumn{1}{c}{\textbf{GPQA}} & \multicolumn{1}{c}{\textbf{MMLU-Pro}} \\ \midrule
\multirow{4}{*}{\begin{tabular}[c]{@{}c@{}}Qwen2.5-\\ Math-7B\end{tabular}}
 & DAPO (teacher) & 37.13 & 15.21 & 69.39 & 40.58 & 38.26 & 43.93 \\
 & OPD & 38.12 & 15.21 & 68.59 & 40.64 & 39.39 & 43.34 \\
 & MOPD & 38.85 & 15.93 & 68.56 & 41.11 & 38.76 & 43.81 \\
 & \textbf{ARMOR} & \textbf{43.04} & \textbf{18.13} & \textbf{76.13} & \textbf{45.77} & \textbf{42.49} & \textbf{46.03} \\ \midrule
\multirow{4}{*}{\begin{tabular}[c]{@{}c@{}}Qwen3-8B-\\ Base\end{tabular}}
 & DAPO (teacher) & 36.98 & 28.13 & 71.72 & 45.61 & 49.87 & 65.55 \\
 & OPD & 38.23 & 27.81 & 71.16 & 45.73 & 52.27 & 65.91 \\
 & MOPD & 37.80 & 29.38 & 72.03 & 46.40 & 50.51 & 65.99 \\
 & \textbf{ARMOR} & \textbf{48.13} & \textbf{34.48} & \textbf{80.35} & \textbf{54.32} & \textbf{56.25} & \textbf{68.90} \\ \bottomrule
\end{tabular}
}
\end{table*}

\section{Implementation Details}\label{appendix:impl}
We implement our method based on the open-source \href{https://github.com/verl-project/verl/tree/v0.5.0/recipe/dapo}{DAPO recipe}. 
Below we detail the specific configurations for reproducibility, and we have also uploaded our codebase to: \href{https://github.com/Hesse73/ARMOR}{https://github.com/Hesse73/ARMOR}.

\textbf{Training Hyperparameters.}
We use a learning rate of $1\text{e-}6$ with a 10-step warmup. 
For the DAPO objective, we adopt the default dual-clip ratios of $\epsilon_\mathrm{low}=0.2$ and $\epsilon_\mathrm{high}=0.28$.
The training process is configured with a global batch size of $B=512$ queries and a response group size of $|G| = 16$.
Optimization is performed with a mini-batch size of 32 prompts, resulting in $\mu = 16$ gradient updates per RL step.
For QAE, we set the baseline quantile to 45\%.

\textbf{ARMOR Configurations.}
For \textbf{Anchor Rollout}, we set a sampling budget of $|\mathcal Y_{\text{ref}|x}| = 4$ per query for the reference policy to conduct rejection sampling, ensuring strict correctness ($R=1$).
Consequently, given the original group size of $|G|=16$, the total inference cost is estimated at approximately $20/16 = 125\%$ of the on-policy baseline.
The mixing coefficient for \textbf{Mixed Optimization} is set to $\alpha = 0.9375$ for Qwen2.5-Math-7B and $\alpha = 0.875$ for Qwen3-8B-Base.

\textbf{Model-Specific Settings.}
To accommodate different context window capabilities, the overlong penalty thresholds (and max generation lengths) are set to 4k (8k) for Qwen2.5-Math-7B and 16k (20k) for Qwen3-8B-Base, leveraging the latter's long-context pretraining \citep{Qwen3}.
Regarding the \textbf{reference reset strategy}, we apply distinct schedules to produce the results in Tab.~\ref{tab:main}, for which we provide a detailed analysis in the following section:
\begin{itemize}[nosep]
    \item \textbf{Qwen2.5-Math-7B:} A fixed reset frequency of $\tau = 160$ for both DAPO and QAE.
    \item \textbf{Qwen3-8B-Base:} A two-stage strategy, starting with $\tau = 1$ for the first 180 steps, followed by a fixed reference policy.
\end{itemize}


\section{Additional Experiment Results}\label{appendix:add_exp}

\begin{figure}[t]
    \centering
    \includegraphics[width=\linewidth]{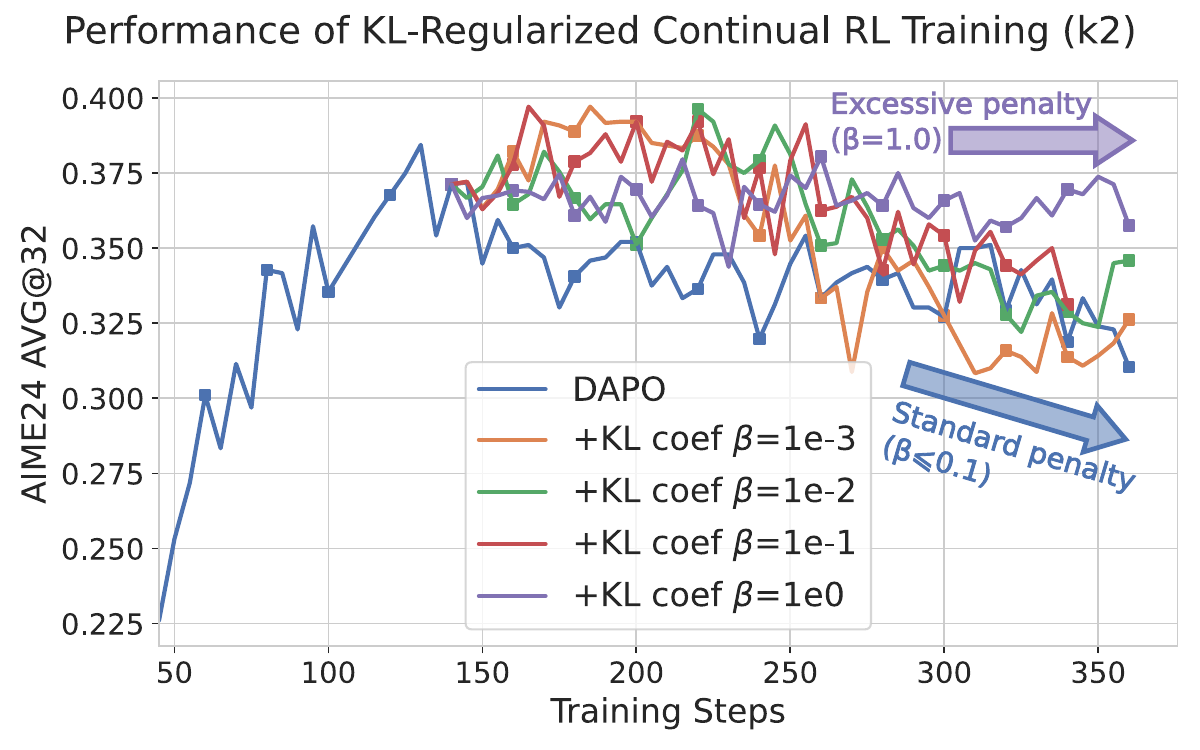}
    \caption{\textbf{KL-regularized training performance (using \texttt{k2} estimation).}
    Consistent with \texttt{k3} results (Fig.~\ref{fig:kl_perf}), standard penalties ($\beta \le 0.1$: \textcolor{myorange}{1e-3}, \textcolor{mygreen}{1e-2}, \textcolor{myred}{1e-1}) merely delay validation collapse. The extreme setting (\textcolor{mypurple}{$\beta$ = 1.0}) enforces stability but induces stagnation, validating the stability-plasticity dilemma in KL-regularization.}
    \label{fig:kl_perf_k2}
\end{figure}

\begin{figure*}[t]
    \centering
    \includegraphics[width=\linewidth]{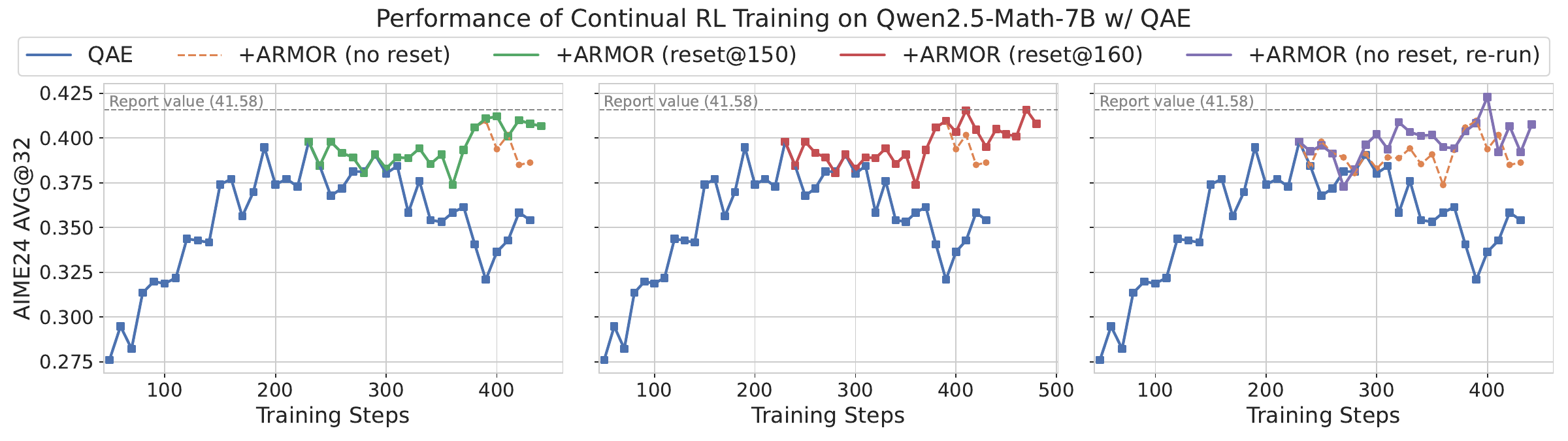}
    \caption{Hyperparameter robustness of ARMOR on QAE.
    We compare ARMOR variants against the \textcolor{myblue}{QAE baseline}.
    Across all settings---whether using \textcolor{myorange}{static anchors}, periodic resets after \textcolor{mygreen}{150 steps}, \textcolor{myred}{160 steps}, or \textcolor{mypurple}{a distinct re-run}---ARMOR consistently stabilizes training and maintains high performance.
    Notably, all variants approach or surpass the reported peak performance (AIME24 \texttt{avg@32=41.58}, indicated by the \textcolor{gray}{gray dashed line}).}
    \label{fig:qae}
\end{figure*}

\subsection{KL Regularization with \texttt{k2} Estimation}
In Section~\ref{sec:kl}, we analyzed continual training using the standard \texttt{k3} estimator for reverse KL.
To ensure our conclusions are robust to estimator choice, we extend this analysis using the \texttt{k2} estimator, which is shown to provide unbiased gradient estimation \citep{klpitfalls}.
As illustrated in Fig.~\ref{fig:kl_perf_k2}, the performance dynamics mirror those observed with \texttt{k3}: standard penalties ($\beta \le 0.1$) fail to prevent eventual collapse, whereas excessive penalties ($\beta=1.0$) secure stability only at the cost of stagnation.
This confirms that the stability-plasticity dilemma is intrinsic to the KL objective itself, rather than an artifact of the estimation method.

\subsection{Comparison with Teacher-Guided Two-Stage Training}\label{appendix:two_stage}

Beyond directly continuing RL from the best zero-RL checkpoint, another natural two-stage strategy is to use this checkpoint as a teacher policy for on-policy distillation.
We therefore compare ARMOR with OPD~\citep{agarwal2024policy} and MOPD~\citep{xiao2026mimo} under the same DAPO-trained reference policy.
As shown in Tab.~\ref{tab:two_stage}, OPD and MOPD provide stable teacher-guided training but only marginally improve over the reference checkpoint.
In contrast, ARMOR achieves substantially larger gains across both base models, indicating that explicitly targeting over-optimization is more effective than simply distilling from the zero-RL model.

\subsection{Hyperparameter Robustness}

While the main body of our paper focuses on ablations using DAPO + Qwen2.5-Math-7B, we further evaluate the robustness of ARMOR's hyperparameters across different algorithms and base models.

Firstly, we present a detailed comparison using the \textbf{QAE + Qwen2.5-Math-7B} setting.
Starting from the best QAE checkpoint (Step 230), we assess four variants of ARMOR to analyze the impact of reset frequency and training stochasticity. 
As illustrated in Fig.~\ref{fig:qae}, we compare:
\begin{enumerate}
    \item \textcolor{myorange}{ARMOR (no reset)}: We freeze the reference policy $\pi_\text{ref}$ throughout the training phase.
    \item \textcolor{mygreen}{ARMOR (reset@150)}: The reference policy is updated ($\pi_\text{ref}\leftarrow\pi_\theta$) every 150 steps during continual training.
    \item \textcolor{myred}{ARMOR (reset@160)}: The reference policy is updated every 160 steps.
    \item \textcolor{mypurple}{ARMOR (no reset, re-run)}: This is a reproducibility check on no-reset. To verify stability against the inherent randomness of RL training (e.g., rollout randomness), we conduct an independent re-run of the static anchor setting.
\end{enumerate}

\begin{figure*}[t]
    \centering
    \includegraphics[width=\linewidth]{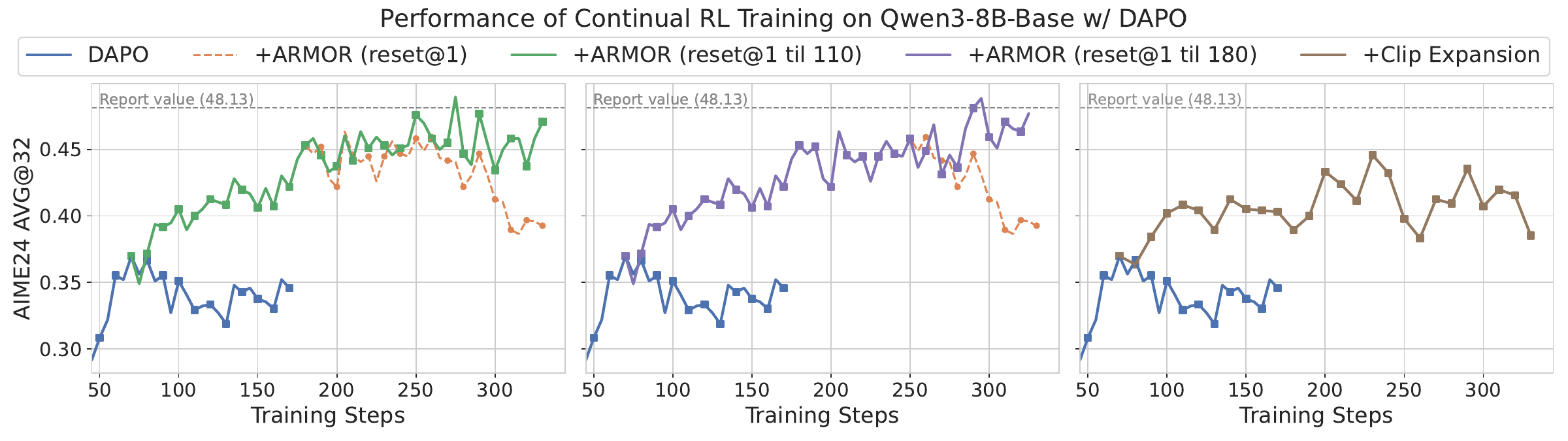}
    \caption{Ablation of Reset Strategy on Qwen3-8B-Base.
    Due to Qwen3's strong capability, continuous reference updates (\textcolor{myorange}{reset@1, \ie $\tau=1$}) enable rapid early gains but suffer from late-stage instability (orange dashed line).
    Our Hybrid Strategy (\textcolor{mygreen}{reset@1 til 110} and \textcolor{mypurple}{reset@1 til 180}) resolves this by freezing the anchor after an initial adaptation phase, both reaching the reported peak performance (AIME24 \texttt{avg@32=48.13}).
    In contrast, the \textcolor{mybrown}{Static Clip Expansion} baseline yields lower and more unstable performance, confirming that ARMOR's benefit stems from structure-aware regularization rather than simple clip relaxation.}
    \label{fig:qwen3}
\end{figure*}

The results in Fig.~\ref{fig:qae} demonstrate strong algorithmic robustness. 
While the \textcolor{myblue}{baseline QAE} suffers from performance degradation, all ARMOR variants successfully sustain the performance and approach or surpass the reported AIME24 value in Tab.~\ref{tab:main}, demonstrating that the gains are robust.

Next, we present a comparison using the \textbf{DAPO + Qwen3-8B-Base} setting.

Unlike the Qwen2.5-Math-7B model, Qwen3 is a strong base model with high intrinsic reasoning potential.
We hypothesize that in the continual training setting, such a model benefits from a more aggressive exploration space in the early stages, without being strictly tethered to a static reference.
To validate this, we compare the following strategies (Fig.~\ref{fig:qwen3}):

\begin{enumerate}
    \item \textcolor{myorange}{ARMOR (reset@1)} We reset the reference policy at every step. This maximizes the adaptivity of the trust region but minimizes the ``anchoring'' effect.
    \item \textcolor{mygreen}{ARMOR (reset@1 til 110)} and \textcolor{mypurple}{ARMOR (reset@1 til 180}): We adopt a two-stage schedule: updating the reference at every step ($\tau=1$) for the first $T$ steps (110 or 180) to facilitate rapid adaptation, then freezing the reference policy to secure stability.
    \item \textcolor{mybrown}{Clip Expansion}: To verify that ARMOR's gain is not merely due to a wider clip range, we test a baseline using standard DAPO with a manually expanded clip range matching ARMOR's theoretical boundary (as detailed in Sec.~\ref{sec:mix_ablation}).
\end{enumerate}

\textbf{Analysis.}
The results reveal three key insights:
\textit{(1) Necessity of Anchors}: Continuous updates ($\tau=1$) accelerate early adaptation but eventually degrade due to the loss of anchor stabilization, confirming the need for a fixed reference in the late stage.
\textit{(2) Effectiveness and Robustness}: The hybrid strategy secures the best of both worlds: rapid early exploration and sustained stability. Notably, both freeze timings (180/250 steps) reach the reported ceiling, demonstrating ARMOR's robustness to the specific schedule.
\textit{(3) Superiority over Simple Relaxation}: The instability of Static Clip Expansion confirms that ARMOR's benefit derives from principled, direction-aware regularization rather than a generic relaxation of constraints.

\section{Potential Risks}
This paper is mainly concerned with algorithmic advances in reinforcement learning for large language models. The proposed method does not introduce new application scenarios, datasets, deployment settings, or model capabilities that would create risks beyond those generally associated with LLM post-training. Therefore, we do not foresee specific potential risks unique to this work, while acknowledging that general LLM-related risks such as misuse, hallucination, and biased generation remain relevant.

\section{Artifacts and Licenses}
The artifacts used in this work, including code, datasets, and model checkpoints, are listed below together with their corresponding licenses. We use these artifacts in accordance with their intended research use and license terms.

\begin{itemize}[left=0em,topsep=0pt,itemsep=0pt]
    \item DAPO's training recipe, including the released code, training scripts, and training data: Apache 2.0 License.
    \item AIME datasets: Apache 2.0 License. These datasets are used only for evaluation.
    \item AMC dataset: Apache 2.0 License. The dataset is used only for evaluation.
    \item GPQA dataset: Creative Commons Attribution 4.0 International (CC BY 4.0) License.
    \item MMLU-Pro dataset: MIT License.
    \item Qwen-2.5-Math model: Apache 2.0 License.
    \item Qwen-3-8B-Base model: Apache 2.0 License.
\end{itemize}

We do not introduce new datasets containing private, sensitive, or personally identifiable information. All external artifacts are publicly available, and our use of them is limited to algorithm development and benchmark evaluation.

\section{The use of LLMs}
We utilize LLMs only to polish some of the language of this paper.
All content was originally drafted by the authors. 
The use of LLMs was restricted to refining some pre-existing text, and any suggested modifications were reviewed by the authors to confirm their accuracy and alignment with the original meaning.

\end{document}